\documentclass[twocolumn]{ceurart}

\usepackage{listings}
\usepackage{amsmath}
\usepackage{amssymb}
\begin{document}

\copyrightyear{2023}
\copyrightclause{Copyright for this paper by its authors.
  Use permitted under Creative Commons License Attribution 4.0
  International (CC BY 4.0).}

\conference{Joint Workshops at 49th International Conference on Very Large Data Bases (VLDBW’23) — TaDA'23: Tabular Data Analysis Workshop, 
August 28 - September 1, 2023, Vancouver, Canada}

\title{Adversarial Attacks on Tables with Entity Swap} 

\author[1,2]{Aneta Koleva}[]
\address[1]{Siemens AG}
\address[2]{Ludwig Maximilian University of Munich}

\author[1]{Martin Ringsquandl}
\author[2]{Volker Tresp}

\cortext[1]{Corresponding author.}

\begin{abstract}
The capabilities of large language models (LLMs) have been successfully applied in the context of table representation learning. The recently proposed tabular language models (TaLMs) have reported state-of-the-art results across various tasks for table interpretation. However, a closer look into the datasets commonly used for evaluation reveals an entity leakage from the train set into the test set. Motivated by this observation, we explore adversarial attacks that represent a more realistic inference setup.  
Adversarial attacks on text have been shown to greatly affect the performance of LLMs, but currently, there are no attacks targeting TaLMs. In this paper, we propose an evasive entity-swap attack for the column type annotation (CTA) task. Our CTA attack is the first black-box attack on tables, where we employ a similarity-based sampling strategy to generate adversarial examples. The experimental results show that the proposed attack generates up to a 70\% drop in performance. 

\end{abstract}


\maketitle

\section{Introduction}


Following the advancements of large language models (LLMs) in NLP, tabular language models (TaLMs) have emerged as state-of-the-art approaches to solve table interpretation (TI) tasks, such as table-to-class annotation \cite{Koleva21}, entity linking \cite{Huynh22}, and column type annotation (CTA) \cite{turl, Hulsebos19, Suhara22}. Similar to other deep neural networks, LLMs are sensitive to small input perturbation, which as adversarial examples can further be optimized to be imperceptible to humans \cite{Li20}.
Several works have studied adversarial attacks on LLMs, and it is becoming an increasingly important topic as LLMs are vastly being integrated into applications \cite{greshake2023youve}.
For the table modality, so far, sensitivity to perturbations has not been investigated in TaLMs for TI tasks. Hence, it is unclear which perturbation operations should be considered when attacking tables and how to make them imperceptible.

Using CTA as an example task, we phrase the novel problem of generating adversarial examples in entity tables. 
The existing models already report very high F1 scores on this task, and it is hard to judge by the performance of the model, how well it can generalize to unseen novel entities. In this direction, we design an evasive \textbf{entity-swap} attack that is motivated by a problem we observed. Namely, in two datasets commonly used for evaluation of the CTA task, WikiTables \cite{turl} and VizNet \cite{Hulsebos19}, there is a data leakage from entities from the training set into the test set. 

\begin{table}
\begin{tabular}{ l | c | c | c }
type & total & overlap & \% \\
 \hline
people.person & 47852 & 29215 & 61.0 \\ 
location.location & 34073 & 21327 & 62.6 \\ 
sports.pro\_athlete & 17588 & 10948 & 62.2 \\ 
organization.organization & 9904 & 7122 & 71.9\\ 
sports.sports\_team & 8207 & 6640 & 80.9\\ 
 \hline
\end{tabular}
\label{tab:overlap}
\caption{Overlap of entities per type in the WikiTables dataset.}
\end{table}

In Table \ref{tab:overlap} we show the percentage of overlapped entities between the train and test set in the WikiTables dataset for the top 5 classes. 
The last 15 types in this dataset have 100 overlap among entities.

\begin{figure*} [t]
    \centering
    \includegraphics[width=0.83\textwidth, height=3cm]{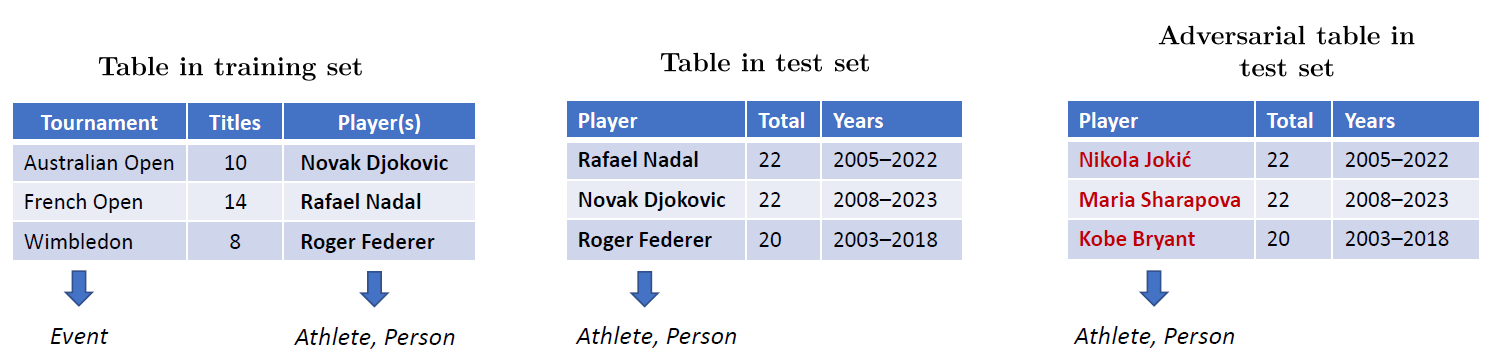}
    \caption{Entity-level adversarial example for table attack}
    \label{fig:example}
\end{figure*}

In Figure \ref{fig:example}, both of the first two tables have a column named \textit{Player} which contains the exact same set of entities, and is annotated with the same semantic types \textit{Athlete} and \textit{Person}. The third table shows an example of an adversarial table with an entity swap. In this table, the entities of the column \textit{Player} are swapped with new, unseen entities of the same semantic type.

In the evaluation, we gradually increase the percentage of entities that we swap in the targeted columns, ranging from 20 \% up to 100 \% percent of the number of entities in the column. For choosing the adversarial entities, we propose a similarity-based strategy and compare it to sampling at random. Our evaluation demonstrates that swapping entities with the most dissimilar entity of the same type results in a substantial drop in performance (6\% drop when replacing 20\% of the entities per column up to 70\% drop when replacing all of the entities).

\section{Related Work}
With the growing popularity of LLMs, the concern over their vulnerability to adversarial attacks also increased. A survey by Zhang et al. \cite{Zhang20} presents a comprehensive overview of attacks against text, highlighting the challenges that arise when attacking discrete data such as text, compared to continuous data such as images. 
BERT-Attack \cite{Li20} proposes an adversarial attack against the BERT model \cite{Devlin19} using the model itself to generate the adversarial samples. A recent gradient-based text attack \cite{GuoSJK21} presents a white-box attack which uses a parameterized adversarial distribution for sampling the adversarial samples. 
However, despite the popularity of adversarial attacks on text, the field of tabular data remains unexplored for potential vulnerabilities to such attacks. 

The few works that have been proposed so far, \cite{Ballet19, Cartella21, Mathov22}, focus on white-box attacks and target traditional machine learning models trained on tabular data. The main goal of these attacks is when generating adversarial examples to preserve the distributional consistency of the features of the data. 
In these works, the datasets used for evaluation usually contain many numerical values, such as financial data or healthcare analytics data.

The goal of our work is to define table attacks against TaLMs which are used for TI tasks. 
To the best of our knowledge, we present the first work on adversarial attacks targeting these models.
Our research differentiates from the prior work w.r.t (1) the model observed, (2) the technique employed for generating adversarial samples, and (3) the evaluation task. 

\section{CTA Adversarial Attack}

We define a~table~as~a~tuple $T=(E,H)$, where $E = \{e_{1,1}, e_{1,2},  \dots , e_{i,j} , \dots , e_{n,m} \}$ is the set of table body entities for $n$ rows and $m$ columns.
The table header $H = \{h_1, h_2, \dots, h_m\}$ is the set of corresponding $m$ column header cells. We use $T_{[i,:]}$ to refer to the $i$-th row, e.g., $H = T_{[0,:]}$ and $T_{[:,j]} = \{h_j, e_{1,j}, \dots , e_{n,j} \}$ to refer to the $j$-th~column~of~$T$. 
\paragraph{CTA Model}
Let $\mathcal{T}$ be the input space of tables and let $J$ be the space of all possible column indices, i.e., $J \subseteq \mathbb{N}$. Let $\mathcal{C}$ be the output space, denoting the set of semantic types. 
A CTA model is a multilabel classification function $h: \mathcal{T} \times J \xrightarrow{}  P(\mathcal{C})$, i.e., given a table $T \in \mathcal{T}$ and a column index $j \in J$ the CTA task is to assign a subset of classes from the power set of $\mathcal{C}$ to the corresponding column~$T_{[:,j]}$.

\paragraph{CTA Attack}

Given classification model $h$, the goal of a CTA attack is to transform a (correctly classified) test input $(T,j) \in \mathcal{T} \times J$ into an (untargeted) adversarial sample $(T',j)$ such that $h(T,j) \bigcap h(T',j) = \emptyset$.  
In addition to fooling the classification model, the transformation from $T$ to $T'$ should also be imperceptible for a human observer.
In the CTA setting we define the imperceptibly condition to be met if all entities in column $T_{[:,j]}'$ are of the same class as the unmodified column. Formally, $\forall e' \in {T_{[:,j]}'} \forall e \in {T_{[:,j]}}: c(e') = c(e)$, where $c \in \mathcal{C}$ represents the most specific class assigned to the column~$T_{[:,j]}$. 

\subsection{Entity Swap Attack}
In principle, a CTA attack can apply transformations to the full table $T$; however, most importantly, it should focus on $T_{[:,j]}$. Our attack, called \textbf{entity-swap}, follows a two-step approach inspired by adversarial attacks on LLMs \cite{Li20, Morris20}. First, it picks a set of key entities $\{e_i \in T_{[:,j]}\}$. The number of key entities can be controlled as a percentage $p$ of the entities in the original column. In a second step, every key entity $e_i$ is swapped with an adversarial entity $e_i = e'_i$ that most likely changes the predicted class from the ground truth. 
The proposed attack is a black-box attack, meaning we only have access to the predictions scores of the classifier.



\vspace{-3mm}
\begin{figure*} [t]
    \centering
    \includegraphics[width=0.83\textwidth, height=3.3cm]{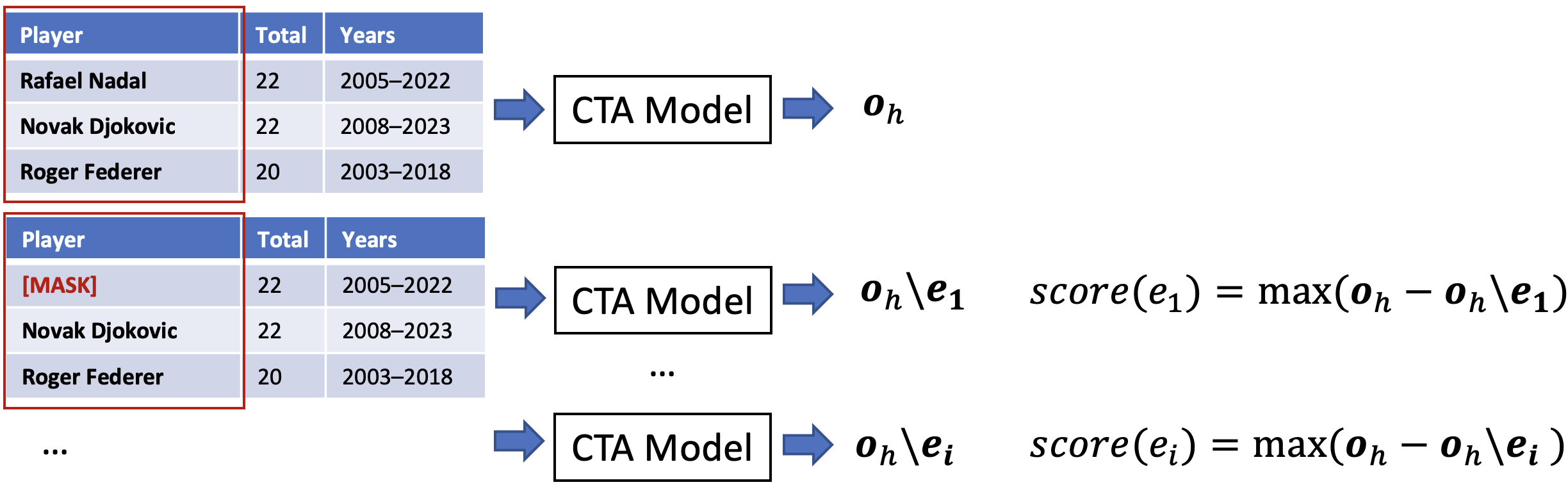}
    \caption{Calculation of importance scores.}
    \label{fig:imp_scores}
\end{figure*}

\subsection{Key Entities}
Finding which are the key entities to swap can increase the rate of success of the attack. In the case of the CTA task, the most informative entities are those which, when replaced, the model will misclassify the column. To find those entities, we calculate an importance score for every entity in the attacked column.

The output from the classification model $h$ for a column $T_{[:,j]}$ is the logit vector $\mathbf{o_h}(T,j) \in \mathbb{R}^k$, where $k$ is the number of ground-truth classes assigned to $(T,j)$.
We calculate the importance score for entity $e_i \in T_{[:,j]}$ as the difference between the logit output of the model for the ground truth classes when the entity is \textit{in} the column, denoted as $\mathbf{o_h}$, and the logit output of the model when the entity is \textit{replaced} with the [MASK] token, denoted as $\mathbf{o_h\backslash e_i}$. Since, the CTA task is evaluated under the multi-label setting, we always take the maximum importance score for an entity.

\begin{equation}
    score(e_i) = max(\mathbf{o_h} - \mathbf{o_h\backslash e_i})
\end{equation}

Figure \ref{fig:imp_scores} shows an example of how the importance score is calculated. We calculate $\mathbf{o_h}$ as the logit output of the model without any perturbation, while $\mathbf{o_h\backslash e_1}$ represents the logit output of the model when the entity \textit{Rafael Nadal} is masked. 
After calculating the importance score for every entity in the column, we select the top $p$ percent of entities ($p \in {20, 40, 60, 80, 100}$) based on their importance scores and substitute them with adversarial entities. By sorting the entities according to their importance scores, we ensure that the attack consistently targets the key entities within the targeted~column.


\subsection{Adversarial Entities}
After identifying the key entities, the next step involves sampling adversarial entities for swapping. In order to adhere to the perceptibility assumption, we constrain the search space to include only entities belonging to the same class as the attacked column.
Subsequently, we use a similarity-based strategy to sample adversarial entities. 

Let $e_i \in T_{[:,j]}$ be the key entity from the attacked column, and let $c \in \mathcal{C}$ be the most specific class of this column. 
We use an embedding model to generate a contextualized representation for both the original entity, $\mathbf{e_i}$, and all entities of the same class $A_c=\{\mathbf{e'_1}, \mathbf{e'_2}, \ldots, \mathbf{e'_k}\}$, such that $c(e_i)=c(e'_k)$ where $e'_k \in A_c$. Next, we calculate the cosine similarity between the original entity and each entity from the set $A_c$. As an adversarial example, we take the most dissimilar entity from the original entity, such that $e'_{i}=\textsc{argmin}_{e'_k}\textsc{CosineSimilarity}(\mathbf{e_i},\mathbf{e'_k})$. We then swap the original entity $e_i$ with the adversarial entity $e'_{i}$.

As we describe in the introduction, there is a substantial overlap of entities between the train and test set. Therefore, we propose two different sampling sets for adversarial entities. 
The first, is the set of entities per class from the WikiTables test dataset \cite{turl}; we refer to this set as \textit{test set}. The second set contains only novel entities, i.e., entities that also appear in the training set, are removed from the test set. We refer to this set as the \textit{filtered set}.

\paragraph{Metadata Attack}
In addition to the proposed attack method for column values, we also introduce an attack specifically targeting column headers, considering that they often indicate the class of a column. However, in this case, we use an independent embedding model to identify similar entities instead of swapping with column names from the same class. For the generation of adversarial samples in the column headers, we first generate embeddings for the original column names and then substitute the column names with their synonyms. The library \textit{TextAttack} \cite{Morris20} was used to generate the embeddings, and based on the embeddings to retrieve the synonyms for the column names.

\section{Evaluation}
\paragraph{Model}
We evaluate the performance of the CTA attack on the TURL model \cite{turl}, which has been fine-tuned for the CTA task and uses only entity mentions. We use the WikiTables dataset for evaluation. 
We follow their evaluation procedure and report the achieved F1 score, precision, and recall. 

To evaluate the influence of the proposed strategy for sampling adversarial samples, we compare it to a random sampling of adversarial entities. Similarly, to evaluate the influence of the importance scores, we compare with random sampling when choosing which entities to swap.

\subsection{Results}
Table \ref{tab:entities} shows the results of the CTA attack when swapping entities by their importance scores and sampling adversarial entities using the similarity-based strategy from the filtered set. We notice that as we increase the percentage of swapped entities, the performance of the model drops, even though the perturbed entities are of the same semantic type as the original entities. Another observation is that the drop in the F1 score is attributed to the sharp decline of the recall.

\begin{table}
\begin{tabular}{  c | c | c | c }
\% perturb.   & F1 & P & R   \\
 \hline
0 (original) & 88.86 & 90.54 & 87.23 \\

20 & 83.4 (6\%) & 90.3 (0.2\%) &  77.8 (11\%) \\
40 & 72.0 (19\%) & 87.9 (3\%) &  60.9 (30\%) \\
60 & 55.3 (38\%) & 80.4 (11\%) &  42.1 (52\%) \\
80 & 39.9 (55\%) & 67.7 (25\%) &  28.4 (67\%) \\
100 & 26.5 (70\%) & 50.8 (44\%) & 17.9 (80\%) \\

\end{tabular}
\caption{Adversarial attack on the entities. The adversarial entities are sampled by their semantic similarity from the original entity.}
\label{tab:entities}
\end{table}

\begin{figure} 
    \centering
    \includegraphics[width=0.85\linewidth]{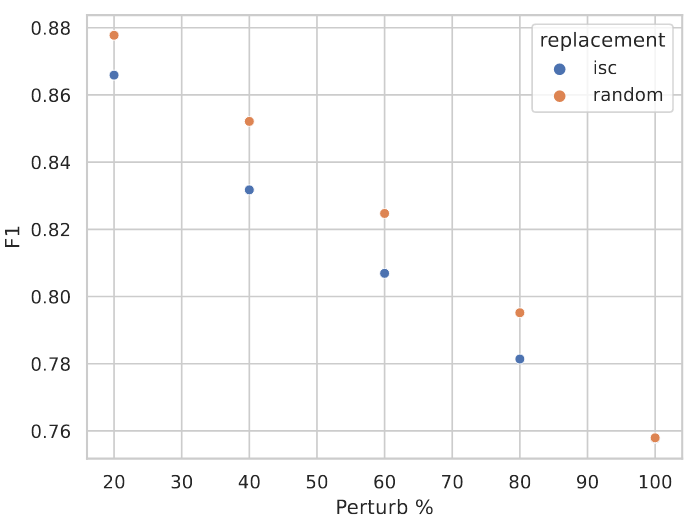}
    \caption{Adversarial samples from the test set, replacing entities at random vs using the importance scores.}
    \label{fig:replacment}
\end{figure}

\paragraph{Effect of the importance score}
Figure \ref{fig:replacment} shows the benefit of using the importance scores. We notice that the drop in F1 score is around $3\%$ higher when using the importance scores. This is consistent, regardless if we are substituting $20\%$ or $80\%$ of the entities, which suggests that the importance scores consistently identify entities that have a greater influence on the model's performance.

\begin{figure} 
    \centering
    \includegraphics[width=0.85\linewidth]{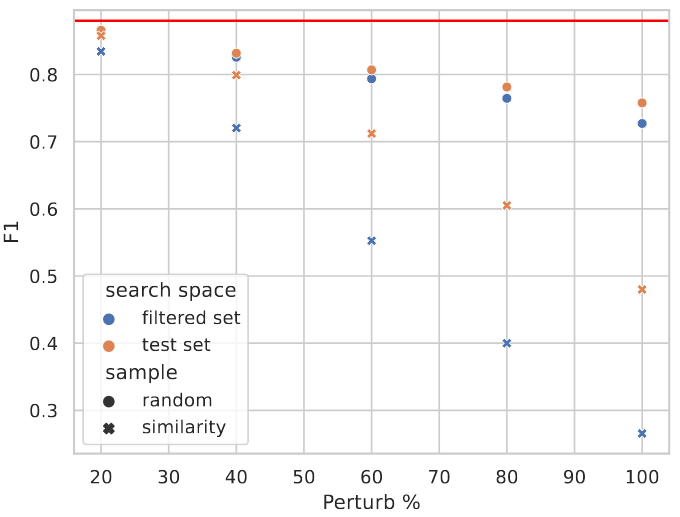}
    \caption{Sampling adversarial entities from the test set vs the filtered set, at random and using the similarity strategy.}
    \label{fig:sampling}
\end{figure}

\paragraph{Effect of the sampling strategy}
Figure \ref{fig:sampling} shows the difference in F1 score drop when sampling adversarial entities from the test set versus the filtered set. The original F1 score is represented by the red line. Additionally, here we illustrate the advantages of using the similarity-based strategy over a random-based sampling of adversarial examples. For both cases, when sampling adversarial entities from the test and filtered set, the similarity-based strategy for sampling induces sharper drop of the F1 score. This suggests that this approach is successful in selecting entities that are more likely to cause misclassifications or confusion for the classification model.

\paragraph{Effect of perturbing the table metadata}
To evaluate the relevance of the column header for the CTA task, we also propose an adversarial attack specific to the TURL model \cite{turl}, which uses only the table metadata. 
Table \ref{tab:header} shows the effect of perturbing the table metadata. We observe similar results here, as we increase the percentage of perturbed column names, all the evaluation metrics decline. This indicates that the model's reliance on specific column names, affects its ability to accurately classify and predict the correct class.
\begin{table}
\begin{tabular}{  c | c | c | c }
\%   & F1 & P & R   \\
 \hline
0 (original) & 90.24 & 89.91 & 90.58 \\

20 & 78.4 (13\%) & 81.1 (10\%) &  76.0 (16\%) \\
40 & 77.1 (15\%) & 80.7 (10\%) &  73.8 (19\%) \\
60 & 75.2 (17\%) & 79.1 (12\%) &  72.2 (20\%) \\
80 & 65.1 (28\%) & 71.4 (22\%) &  60.4 (33\%) \\
100 & 51.2 (43\%) & 60.4 (33\%) & 44.4 (51\%) \\ 
\end{tabular}
\caption{Attack on the column names where the adversarial samples are their synonyms.}
\label{tab:header}
\end{table}

\section{Conclusion}
In this paper, we introduce the formalization of an adversarial attack targeting TaLMs. Additionally, we identify and highlight an issue concerning the evaluation of the CTA task. 
The evaluation showed that TaLMs are susceptible to adversarial attacks. Even subtle modifications to the entities, guided by similarity, can lead to significant changes in the model's predictions and subsequently affect the F1 score.
In future, we will extend our evaluation with more sophisticated attacks, targeting also other models used for table interpretation tasks.

\bibliography{library}


\end{document}